\documentclass[conference,onecolumn]{IEEEtran}
\IEEEoverridecommandlockouts

\usepackage{cite}
\usepackage{amsmath,amssymb,amsfonts}
\usepackage{algorithmic}
\usepackage{graphicx}
\usepackage{textcomp}
\usepackage{xcolor}

\usepackage{hyperref}
\usepackage{cleveref}
\usepackage{multirow}

\newcommand{\inlineheading}[1]{%
  \vspace{0.3\baselineskip}%
  \noindent\textbf{#1.}\hspace{0.5em}%
}

\def\BibTeX{{\rm B\kern-.05em{\sc i\kern-.025em b}\kern-.08em
    T\kern-.1667em\lower.7ex\hbox{E}\kern-.125emX}}
\begin{document}
\bstctlcite{IEEEexample:BSTcontrol}

\title{Diffusion-Based Multi-Class Normality for OOD Detection: An Application to CDP Authentication\\
}

\author{
\IEEEauthorblockN{Bolutife Atoki, Iuliia Tkachenko, 
Bertrand Kerautret, and Carlos Crispim-Junior}
\IEEEauthorblockA{\textit{Université Lumière Lyon 2, CNRS, INSA Lyon, Universite Claude Bernard Lyon 1}\\ \textit{LIRIS, UMR 5205} Lyon, France \\
\{bolutife.atoki, iuliia.tkachenko, bertrand.kerautret, carlos.crispim-junior\}@liris.cnrs.fr}
}

\maketitle

\begin{abstract}
Reconstruction-based generative models offer a natural
framework for unsupervised out-of-distribution (OOD) detection, but
multi-class normality modelling requires a single detector to capture multiple in-distribution manifolds and produce comparable anomaly scores across classes. We study this problem in copy detection pattern (CDP) authentication, where authentic and counterfeit samples are visually similar but differ in subtle printing-and-digitisation (P\&D) signatures. We propose a diffusion-based multi-class normality framework in which a
single class-conditional ControlNet is trained exclusively on authentic CDPs from multiple P\&D classes and detects counterfeits through reconstruction error under authentic-class conditioning. We further introduce dual template masking, which hides complementary regions of the input template and scores only withheld pixels, reducing reliance on visible binary structure. On the Indigo $1\times1$ Base dataset, the proposed method outperforms traditional and adapted generative baselines under multi-class authentic-versus-counterfeit evaluation, without using
counterfeit samples for training or threshold calibration.
\end{abstract}


\section{Introduction}
\label{sec:intro}
The detection of out-of-distribution (OOD) samples (inputs that do not belong to the distribution seen during training) is a fundamental challenge in deploying reliable machine learning systems~\cite{DBLP:conf/iclr/HendrycksG17, DBLP:conf/nips/LiuWOL20}. Reconstruction-based approaches, where generative models are trained on in-distribution data and OOD samples are identified via elevated reconstruction errors, have proven particularly effective for unsupervised anomaly and novelty detection~\cite{DBLP:conf/cvpr/GrahamPTNOC23, DBLP:conf/icml/LiuZWW23}. A natural extension is multi-class normality modelling, where a single model learns the manifolds of multiple in-distribution classes and identifies deviations from any of them. In this work, we apply this paradigm to the authentication of Copy Detection Patterns (CDPs)\footnote{\url{https://en.wikipedia.org/wiki/Copy_detection_pattern}}. CDPs are utilised for product authentication due to their low cost and inherent sensitivity to copying. However, recent advances in deep learning have demonstrated that machine learning-based attacks can accurately estimate CDP templates from printed samples and reproduce high-fidelity counterfeits through reprinting. These DNN-generated counterfeits exhibit remarkable visual similarity to authentic CDPs, challenging traditional authentication methods that rely on image-level similarity comparison. This vulnerability necessitates authentication approaches capable of detecting sophisticated counterfeits despite their similarity to authentic counterparts~\cite{Chaban2021MachineLA, DBLP:conf/wifs/YadavTTF19}.

\begin{figure*}[t]
    \centering
    \includegraphics[width=1.0\linewidth]{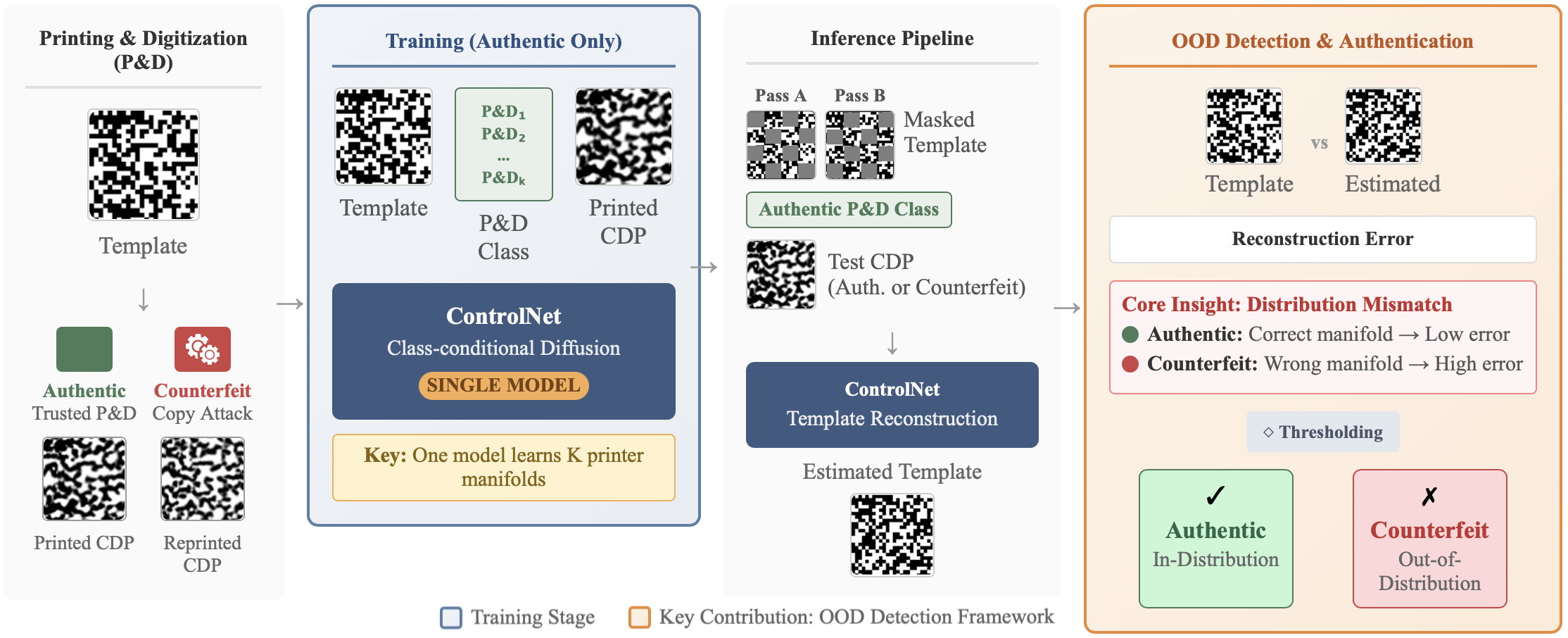}
    \caption{Overview of the proposed framework. A class-conditional ControlNet learns authentic P\&D manifolds and detects counterfeits via dual-mask reconstruction error under authentic-class conditioning. Images are cropped and zoomed for illustration.}
    \label{fig:graphical_abstract}
\end{figure*}

From an OOD detection perspective, this application poses two core challenges that generalize beyond CDPs. First, in-distribution and OOD samples may be structurally near-identical, differing only in subtle distributional signatures. In this case, authentic and counterfeit CDPs share the same nominal template but undergo different printing-and-digitisation (P\&D) processes. Authentic samples are produced using trusted printing and scanning devices, whereas counterfeits result from unauthorised reprinting after template estimation (see ``Printing and Digitisation'' module of \Cref{fig:graphical_abstract}). While these P\&D differences leave subtle, device-specific signatures arising from hardware imperfections and mechanical variations \cite{8726322, 5537996}, exploiting them for authentication is non-trivial due to their spatial non-uniformity and complexity. A second challenge is that the normality model should capture multiple in-distribution classes within a single framework. Practical deployment of CDP authentication methods requires handling multiple printing devices within a single authentication system, as production environments typically span multiple manufacturing sites with different P\&D equipment.

Various authentication approaches have been proposed, spanning analytical modelling, learned representations, and generative synthesis. Early analytical approaches~\cite{DBLP:conf/wifs/TuttTCPBHV22, DBLP:journals/tifs/TuttTCPBHV24, DBLP:conf/wifs/TuttV24} model P\&D processes through statistical codebooks, while learning-based methods such as Siamese networks~\cite{DBLP:conf/wifs/ZeghidiCT23} directly learn printer-specific signatures via feature space embeddings. However, analytical methods assume spatial uniformity that does not hold for real devices, and learned embeddings may struggle to localize device-specific artifacts without template context.

Generative approaches explicitly model P\&D transformations, employing template estimation~\cite{DBLP:conf/wifs/TaranTHCBV21} or printed CDP synthesis~\cite{DBLP:conf/wifs/ChabanPV24, DBLP:conf/wifs/PulferBTCTHV22} with deep neural network–based architectures such as Pix2Pix and U-Net. Diffusion models have been explored for CDP generation~\cite{DBLP:conf/wifs/BelousovTKV23}, though primarily for augmentation. Despite their sophistication, existing generative methods share a critical scalability limitation, as each requires training separate models per printer type, making them impractical for production environments with multiple devices. Additionally, supervised variants~\cite{DBLP:journals/tifs/TuttTCPBHV24} require counterfeit samples during training, which may be unavailable and cannot anticipate evolving attacks. These limitations motivate reconceptualizing CDP authentication as unsupervised out-of-distribution detection, in which the model is trained and calibrated exclusively on authentic data, and counterfeits are identified purely as deviations from learned in-distribution manifolds. Chapus et al.~\cite{DBLP:conf/avss/ChapusCEB25} adopt this perspective, training an energy-based model exclusively on authentic samples for both model fitting and threshold calibration.

Recent work has shown that diffusion models can capture printer
signatures under textual and spatial conditioning
\cite{DBLP:conf/wacv/AtokiTKC26}. Building on that capability, we reformulate
authentication as out-of-distribution detection via multi-class
normality modelling, rather than supervised printer classification. A single model learns distinct authentic manifolds for different authentic classes, and samples outside these manifolds yield elevated reconstruction errors under authentic-class conditioning. Since full-template
access can make this error depend on binary structure rather than subtle P\&D signatures, we introduce dual template masking, following diffusion-inpainting OOD detection~\cite{DBLP:conf/icml/LiuZWW23}, which hides complementary template regions in two reconstruction passes and scores only the withheld pixels.

\begin{figure}[t]
    \centering
    \includegraphics[width=0.5\linewidth]{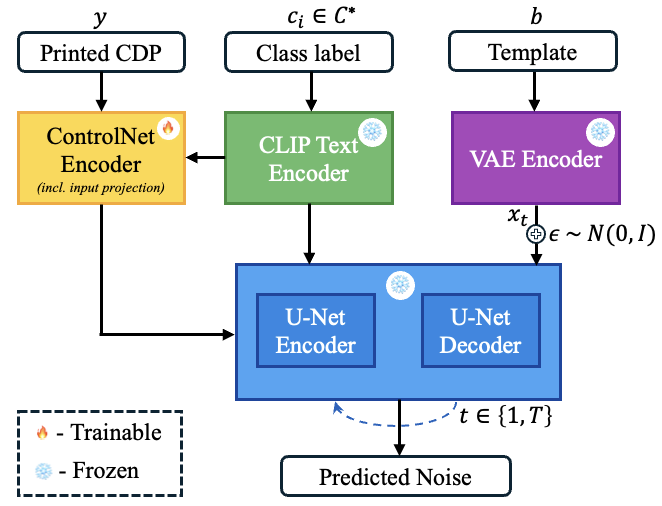}
    \caption{Training architecture. The printed CDP is fed to the ControlNet encoder as a spatial conditioning signal alongside the class prompt via the CLIP text encoder. The VAE encoder maps the binary template to the latent space, where the U-Net denoises under joint conditioning. At each timestep $t$, the network predicts the added noise $\epsilon$.}
    \label{fig:controlnet_training}
\end{figure}

Accordingly, we train a class-conditional ControlNet~\cite{DBLP:conf/iccv/ZhangRA23}
exclusively on authentic CDPs from $K$ P\&D classes. Authentication is performed by
the resulting dual-mask reconstruction error, calibrated solely from authentic
validation samples (see \Cref{fig:controlnet_training}).

The main contributions of this paper are as follows. We propose a diffusion-based multi-class normality framework for unsupervised OOD detection, in which a single class-conditional model learns multiple in-distribution manifolds and identifies OOD samples via conditioning mismatch and reconstruction error analysis. We demonstrate the effectiveness of this framework by formulating CDP authentication as a class-conditional OOD detection problem. In addition, we introduce dual template masking to hide complementary
regions of the input template and score only withheld pixels, reducing
reliance on visible binary structure and improving sensitivity to subtle
class-specific P\&D deviations. Finally, we adapt prior work as baselines for multi-class evaluation. Our implementation and models are available at our public repository \footnote{Repository URL is available at \url{https://gitlab.liris.cnrs.fr/anr-trustit/cdp-multiclass-normality.git}}

The remainder of this paper is organized as follows. \Cref{sec:methodology} details our methodology, \Cref{sec:experiments} describes the experimental setup and baselines, \Cref{sec:results-and-discussion} discusses results, and \Cref{sec:conclusion} concludes with a summary of our findings, limitations, and future directions.

\section{Methodology}
\label{sec:methodology}

\subsection{Problem Formulation and Overview}
\label{subsec:problem-formulation}
We formulate multi-class normality modelling as a class-conditioned diffusion process for OOD detection in the context of CDP authentication. Let $b \in \{0,1\}^{H \times W}$ denote the binary template, $p$ the authentic CDP printed from $b$ using a trusted P\&D device from one of $K$ classes $\{c_1, \ldots, c_K\}$, and $\tilde{p}$ a counterfeit produced from $p$. The authentication task is: given a candidate CDP $y \in \{p, \tilde{p}\}$, determine whether it belongs to the multi-class authentic distribution or represents an out-of-distribution counterfeit. At inference, the binary template $b$ is available (stored and linked to the product), and a conditioning class $c_i$ is inferred from $y$ over the authentic classes, as formalised in \Cref{subsec:authentication}. We now describe the architectural components underlying this reconstruction-based detection framework.

\subsection{Class-Conditional Diffusion for Multi-Class Normality Modelling}
\label{subsec:Multi-Class-Normality-Modelling}
Our class-conditional diffusion backbone builds on the ControlNet-based
CDP authentication framework of~\cite{DBLP:conf/wacv/AtokiTKC26}, which adapts
latent diffusion~\cite{DBLP:conf/cvpr/RombachBLEO22} to jointly condition
on the printed CDP and a textual printer prompt. We repurpose this
architecture for authentic-only multi-class normality modelling, using
reconstruction error for unsupervised OOD detection. Let $x_0$ denote the latent
representation of the binary template $b$, let $z$ denote the latent
representation of the printed CDP, and let $c_i$ denote the printer
prompt. The denoising network $\epsilon_\theta(x_t, t, z, c_i)$ predicts
the noise $\epsilon$ added at timestep $t$, and training minimizes:
\begin{equation}
\mathcal{L} = \mathbb{E}_{x_0, \epsilon \sim \mathcal{N}(0,I), t, z, c_i} \left[ \|\epsilon - \epsilon_\theta(x_t, t, z, c_i)\|_2^2 \right]
\label{eq:diffusion_loss}
\end{equation}

where the expectation $\mathbb{E}$ is taken over all training samples across $K$ authentic P\&D classes, and $\mathcal{N}(0, I)$ denotes the standard normal distribution.

\subsection{Dual-Mask Template Masking}
\label{subsec:template-masking}

The full input template provides a strong structural cue for
reconstruction. If used directly, the resulting error may depend on
reproducing visible binary structure rather than on subtle class-specific
P\&D signatures. Following the masking principle of diffusion-inpainting
OOD detection~\cite{DBLP:conf/icml/LiuZWW23}, we therefore evaluate
reconstruction only on template regions hidden from the model, so that
the score reflects how well missing structure can be inferred under the
authentic-class conditioning.

We use a checkerboard mask $M \in \{0,1\}^{H \times W}$, where
$M_{ij}=1$ denotes pixels visible in pass $A$ and $M_{ij}=0$ denotes
pixels hidden in pass $A$. Let $\bar{M}=\mathbf{1}-M$ be the inverted
mask used for pass $B$. Hidden pixels are replaced by the constant value
$-1$:
\begin{align}
b_{\mathrm{masked}}^{(A)}
&= M \odot b + (\mathbf{1}-M)(-1),
\label{eq:dual_mask_input_A}\\
b_{\mathrm{masked}}^{(B)}
&= \bar{M} \odot b + (\mathbf{1}-\bar{M})(-1).
\label{eq:dual_mask_input_B}
\end{align}

Let $b_{\mathrm{rec}}^{(A)}$ and $b_{\mathrm{rec}}^{(B)}$ denote the
corresponding reconstructions of the masked versions of binary template $b$. Since pass $A$ hides the pixels in
$\bar{M}$ and pass $B$ hides the pixels in $M$, we compute each error
only on the pixels hidden in that pass:
\begin{equation}
e_{\mathrm{dual}}
=
\frac{1}{2}
\left[
e\!\left(b_{\bar{M}}, b_{\mathrm{rec},\bar{M}}^{(A)}\right)
+
e\!\left(b_M, b_{\mathrm{rec},M}^{(B)}\right)
\right],
\label{eq:dual_mask_error}
\end{equation}
where $b_M$ and $b_{\bar M}$ denote template values restricted to the
corresponding mask supports, and $e(\cdot,\cdot)$ is computed only over
the selected pixels.

\subsection{Authentication via OOD Detection}
\label{subsec:authentication}
Given a candidate CDP $y$, the binary template $b$, and conditioning
class $c_i$ (inferred as described below), we apply the complementary
dual-mask strategy (\Cref{subsec:template-masking}) to produce
$e_{\text{dual}}$, thresholded against $\tau$ calibrated from
authentic validation samples:

\begin{equation}
\text{Auth}(y) =
\begin{cases}
\text{Authentic,} & \text{if } e_{\text{dual}} \leq \tau \\
\text{Counterfeit,} & \text{otherwise}
\end{cases}
\label{eq:auth_decision}
\end{equation}

At test time, the conditioning prompt is inferred using the
diffusion-classifier procedure of~\cite{DBLP:conf/iccv/LiPDBP23,DBLP:conf/wacv/AtokiTKC26}.
For each authentic class prompt $c_i \in \mathcal{C}^*$, the model
evaluates the expected noise-prediction error and selects the minimiser:

\begin{equation}
\hat{c} = \arg\min_{c_i \in \mathcal{C}^*} \; \mathbb{E}_t \left[
\bigl\| \epsilon - \epsilon_\theta(x_t,\, t,\, z,\, c_i) \bigr\|_2^2 \right]
\label{eq:self_class}
\end{equation}

where $\mathcal{C}^* = \{c_1, \ldots, c_K\}$ is the set of authentic
P\&D classes exclusively, and $\mathbb{E}_t$ is estimated over $N$
randomly sampled timesteps. Unlike~\cite{DBLP:conf/wacv/AtokiTKC26}, the
search is restricted to authentic classes only, yielding test-time
inferred conditioning without assuming access to counterfeit
identities. This remains informative because a counterfeit
$\tilde{p}$ derived from an authentic print of device $c_j$ retains
residual source-family P\&D signatures, so the inferred prompt is often
family-consistent even when the sample lies outside the authentic
manifold. The subsequent reprinting step nevertheless perturbs that
manifold sufficiently that $e_{\text{dual}}$ remains high under
$\hat{c}$, enabling OOD detection.

\section{Experiments}
\label{sec:experiments}
\begin{table*}[ht]
\centering
\caption{Authentication performance at the fixed global threshold ($\lambda=1\sigma$) under inferred conditioning. Lower is better.}
\label{tab:acc-method-comparison}
\tiny
\resizebox{\textwidth}{!}{%
\begin{tabular}{llccccccccc}
\hline
                          &                              & \multicolumn{1}{l}{$\textbf{P}_{\text{err}}$} & \multicolumn{3}{c}{$\textbf{P}_{\text{miss}}$}                                 & \multicolumn{5}{c}{$\textbf{P}_{\text{fa}}$}                                                                                  \\[0.2em] \hline
                          & \multicolumn{1}{l|}{}        & \multicolumn{1}{l|}{}             & \textit{HP55} & \textit{HP76} & \multicolumn{1}{c|}{\textit{Mean}} & \multicolumn{1}{l}{\textit{HP55\_55}} & \textit{HP55\_76} & \textit{HP76\_76} & \textit{HP76\_55} & \textit{Mean} \\
\multirow{5}{*}{Baseline} & \multicolumn{1}{l|}{Chaban et al.~\cite{DBLP:conf/wifs/ChabanPV24}}             & \multicolumn{1}{c|}{0.106}           & 0.049                 & 0.250            & \multicolumn{1}{c|}{0.149}             & 0.097                & 0.153                & \textbf{0.000}       & 0.000                & 0.063            \\
                          & \multicolumn{1}{l|}{Taran et al.~\cite{DBLP:conf/wifs/TaranTHCBV21}}              & \multicolumn{1}{c|}{0.130}           & \textbf{0.083}        & \textbf{0.007}   & \multicolumn{1}{c|}{\textbf{0.045}}    & 0.333                & 0.510                & 0.021                & 0.000                & 0.216            \\
                          & \multicolumn{1}{l|}{Tutt et al.~\cite{DBLP:journals/tifs/TuttTCPBHV24}}   & \multicolumn{1}{c|}{0.238}  & 0.007       & 0.417  & \multicolumn{1}{c|}{0.212}               & 0.542                  & 0.521                  & 0.000                  & 0.000                  & 0.266              \\
                          & \multicolumn{1}{l|}{NCC}                       & \multicolumn{1}{c|}{0.286}           & 0.292                 & 0.310            & \multicolumn{1}{c|}{ 0.301}            & 0.300                & 0.361                & 0.264                & 0.201                &  0.273            \\
                          & \multicolumn{1}{l|}{Chapus et al.~\cite{DBLP:conf/avss/ChapusCEB25}}  & \multicolumn{1}{c|}{0.335}  & 0.000       & 0.340  & \multicolumn{1}{c|}{0.170}               & 1.000                  & 1.000                  & 0.000                  & 0.000                  & 0.500              \\ \hline
\multirow{3}{*}{Ours}     & \multicolumn{1}{l|}{MSE}     & \multicolumn{1}{c|}{\textbf{0.055}}  & 0.180            & 0.021            & \multicolumn{1}{c|}{0.101}            & \textbf{0.000}                          & \textbf{0.000}      & 0.042                & \textbf{0.000}      & \textbf{0.010}            \\
                          & \multicolumn{1}{l|}{PCC}     & \multicolumn{1}{c|}{0.061}           & 0.194            & 0.021            & \multicolumn{1}{c|}{0.108}            & 0.000                                    & 0.007                & 0.042                & 0.007                & 0.014            \\
                          & \multicolumn{1}{l|}{BER}     & \multicolumn{1}{c|}{0.063}           & 0.201            & 0.021            & \multicolumn{1}{c|}{0.111}            & 0.000                                    & 0.007                & 0.042                & 0.014                & 0.016            \\ \hline
\end{tabular}
}
\end{table*}
\subsection{Experimental Setup}
\label{subsec:experimental-setup}
\inlineheading{Dataset} The Indigo $1\times1$ Base dataset~\cite{Chaban2021MachineLA} contains 720 binary templates, each printed by two authentic printing devices (HP Indigo 5500 and HP Indigo 7600, denoted HP55 and HP76) and four counterfeit types (HP55\_55, HP55\_76, HP76\_55, HP76\_76), yielding 4,320 samples. Counterfeits are generated by estimating templates from authentic prints and reprinting them with each device.
To support out-of-distribution (OOD) detection, we reformulate the dataset as a multi-class problem. Specifically, template–print pairs from the two authentic devices ($K=2$) are treated as in-distribution (P\&D) classes, while the four counterfeit types are used exclusively for OOD evaluation.
We construct training, validation, and test splits of 70\%, 10\%, and 20\%, respectively, based on template identity (504/72/144 samples per class). This protocol prevents leakage across splits~\cite{DBLP:conf/wifs/TaranTHCBV21, DBLP:conf/wifs/ChabanPV24, DBLP:journals/tifs/TuttTCPBHV24} and ensures evaluation on entirely unseen templates.

\inlineheading{Detection scores and thresholding}Reconstruction error is measured using MSE, BER, and PCC~\cite{DBLP:conf/wifs/BelousovPCTTHV22, Chaban2021MachineLA, DBLP:conf/wifs/YadavTTF19}. For each metric \(m\), a detection threshold is estimated from authentic validation samples only:
\begin{equation}
\tau^m = \mu^m + s_m \lambda \sigma^m
\label{eq:threshold}
\end{equation}
where \(\mu^m\) and \(\sigma^m\) denote the mean and standard deviation of \(m\) on the authentic validation set, \(\lambda\) controls detection sensitivity, and \(s_m \in \{+1,-1\}\) determines the threshold direction: \(s_m=+1\) for distance-based metrics, for which larger values indicate greater deviation from authentic samples, and \(s_m=-1\) for similarity-based metrics, for which smaller values indicate greater deviation.

\inlineheading{Evaluation protocol} Results are reported under two scenarios. In the fixed-threshold scenario, $\lambda=1$ is held constant across all methods. We report in \Cref{tab:acc-method-comparison}, the missed authentic detection rate $P_{\text{miss}}$, false-acceptance rate $P_{\text{fa}}$, and balanced error rate: \begin{equation} 
P_{\text{err}} = \frac{P_{\text{miss}} + P_{\text{fa}}}{2} 
\label{eq:auth-eval} 
\end{equation}

In the threshold-robustness scenario, the decision threshold is swept over the full range of reconstruction error scores to obtain the ROC curve, and AUROC is reported, aggregating all authentic classes against all counterfeit types (\Cref{fig:roc-curves}).

\subsection{Implementation Details}
\label{subsec:implementation-details}

We adopt the data augmentation and optimization settings
from~\cite{DBLP:conf/wacv/AtokiTKC26}. Specifically, each template--printed CDP
pair is augmented 20 times using random crops and flips, while printed
CDPs additionally receive photometric perturbations to model imaging
variability and templates remain binary. The VAE is fine-tuned for CDP
images, while the U-Net backbone and CLIP encoder are frozen; training
uses mixed precision on an RTX 3090, batch size 8 with gradient
accumulation of 4, AdamW optimization, a cosine schedule with 500-step
warmup, learning rate $8 \times 10^{-5}$, and 200 epochs. The
task-specific differences are the authentic-only multi-class training
protocol, test-time prompt inference over authentic classes, global
thresholding with $\lambda=1.0$ from \Cref{eq:threshold}, and the
dual-mask template masking with $8 \times 8$ grids at
inference.

\subsection{Baseline Methods}
\label{subsec:baseline-methods}
We compare against NCC, Pix2Pix-based print synthesis of Chaban et al.~\cite{DBLP:conf/wifs/ChabanPV24},
OC-SVM on $D_{tt}/D_{xx}$ features from Taran et al.~\cite{DBLP:conf/wifs/TaranTHCBV21}, the analytical LLS score of Tutt et al.~\cite{DBLP:journals/tifs/TuttTCPBHV24} and the authentic-only energy-based model of Chapus et al.~\cite{DBLP:conf/avss/ChapusCEB25}. Each method is reduced to a scalar anomaly or similarity score and evaluated as a single authentic-versus-counterfeit detector across all authentic P\&D classes, with thresholds calibrated only on authentic validation data. For similarity scores such as NCC and PCC, the threshold direction is reversed so that lower similarity indicates greater anomaly.

\section{Results and Discussion}
\label{sec:results-and-discussion}
The evaluation addresses two related but distinct requirements. First,
\emph{per-family authentic-versus-counterfeit discrimination} measures
whether a method separates authentic samples from counterfeits associated
with the same source printer family, e.g., HP55 authentic samples versus
HP55-derived counterfeits. This evaluates whether the score captures
family-specific P\&D deviations. Second, \emph{multi-class
authentication} requires a single detector to operate across all
authentic P\&D classes using one common score scale and one global
threshold. This additionally tests cross-class score calibration, where authentic samples from different printers should receive comparable normality scores, while counterfeits from all families should be shifted toward the anomalous side.

\begin{figure}[t]
    \centering
    \includegraphics[width=0.7\linewidth]{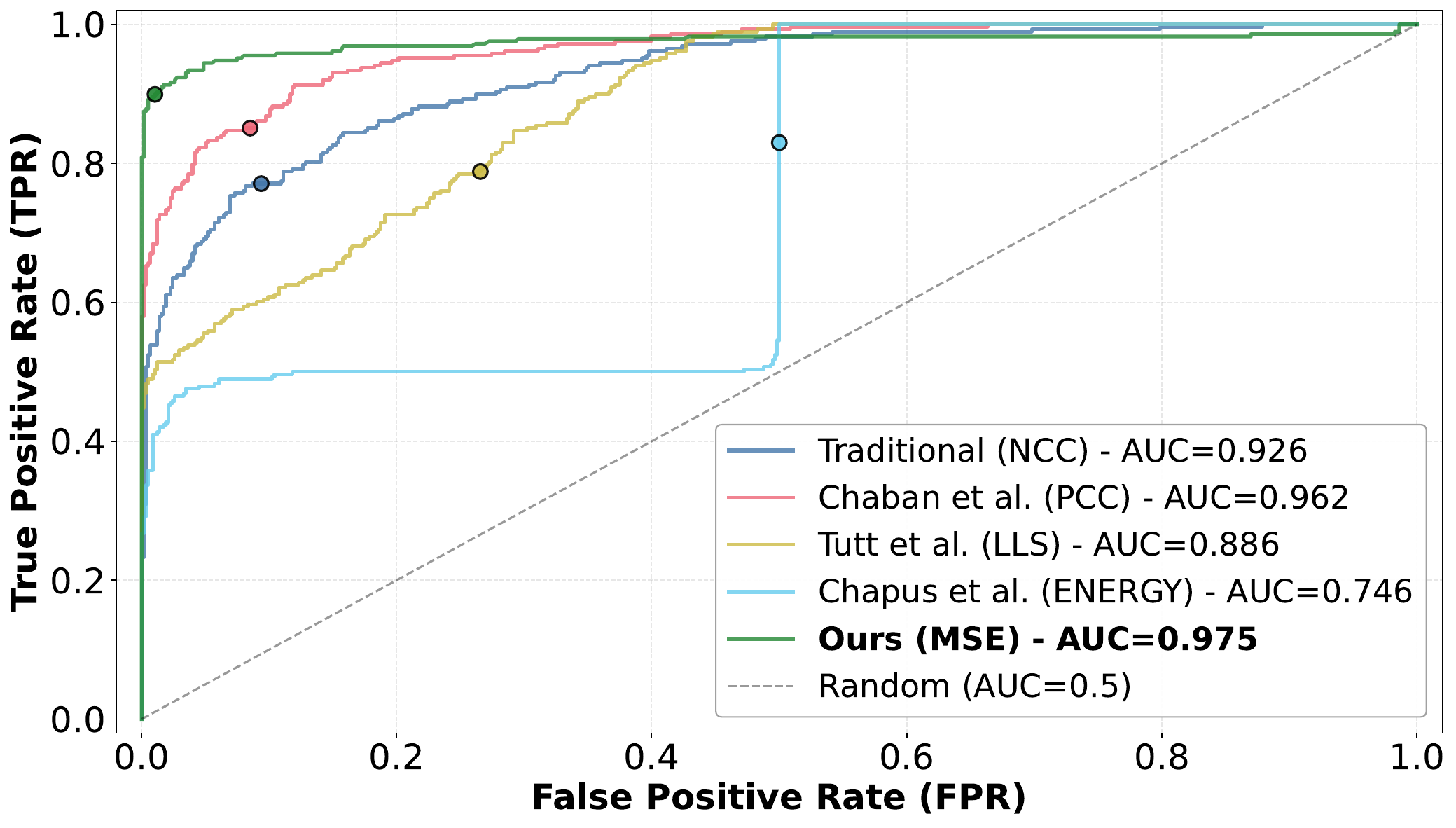}
    \caption{ROC curves for multi-class authentication. All authentic classes are evaluated against all counterfeit classes using a single score per method. Dots mark the $\lambda=1\sigma$ operating points from \Cref{tab:acc-method-comparison}. Baselines are adapted.}
    \label{fig:roc-curves}
\end{figure}

\inlineheading{Multi-class authentication performance}The proposed method achieves the strongest performance in the multi-class
authentication setting. With MSE scoring, it obtains the lowest
$P_{\mathrm{err}}=0.055$ and the highest combined AUROC of $0.975$,
compared with $P_{\mathrm{err}}=0.106$ and AUROC $=0.962$ for the best
adapted baseline, Chaban et al.~\cite{DBLP:conf/wifs/ChabanPV24}, as shown in
\Cref{tab:acc-method-comparison} and \Cref{fig:roc-curves}. This indicates that the dual-mask
reconstruction score is not only discriminative, but also sufficiently
aligned across authentic P\&D classes to support a single global decision
rule. The same trend holds under template-level bootstrap resampling:
\Cref{tab:combined-bootstrap-ci} shows that our $P_{\mathrm{err}}$ confidence
interval does not overlap that of Chaban et al.~\cite{DBLP:conf/wifs/ChabanPV24} while the AUROC
intervals overlap only slightly.

\begin{figure}[t]
    \centering
    \includegraphics[width=1.0\linewidth]{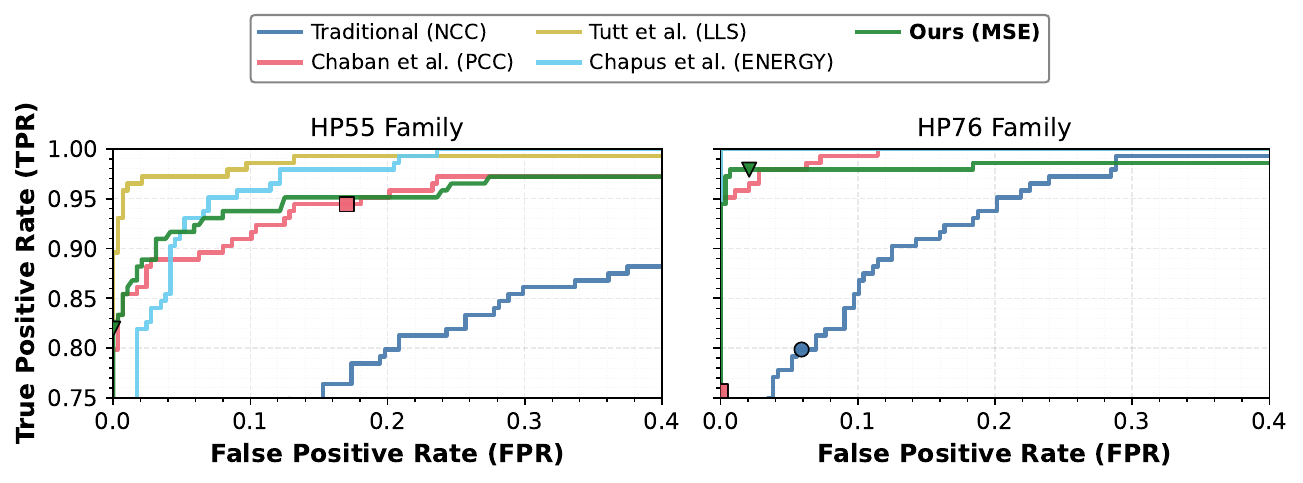}
    \caption{Per-family ROC curves. Each panel evaluates authentic samples and counterfeits associated with one source printer family separately, therefore measuring discrimination without requiring score calibration across authentic classes. Dots mark the fixed-threshold operating points at $\lambda=1\sigma$.}
    \label{fig:roc-perfamily}
\end{figure}

\inlineheading{Per-family discrimination versus multi-class calibration}In \Cref{fig:roc-perfamily}, we see that several baselines remain effective when
each source printer family is evaluated separately. In this setting,
methods are not required to place HP55 and HP76 authentic samples on the same score scale, rather, they only need to rank authentic and counterfeit samples correctly within a given family. For example, Tutt et al.~\cite{DBLP:journals/tifs/TuttTCPBHV24} and Chapus et al.~\cite{DBLP:conf/avss/ChapusCEB25} achieve strong per-family AUROCs, showing that their scores contain useful P\&D-discriminative information.

\begin{figure}[t]
    \centering
    \includegraphics[width=0.8\linewidth]{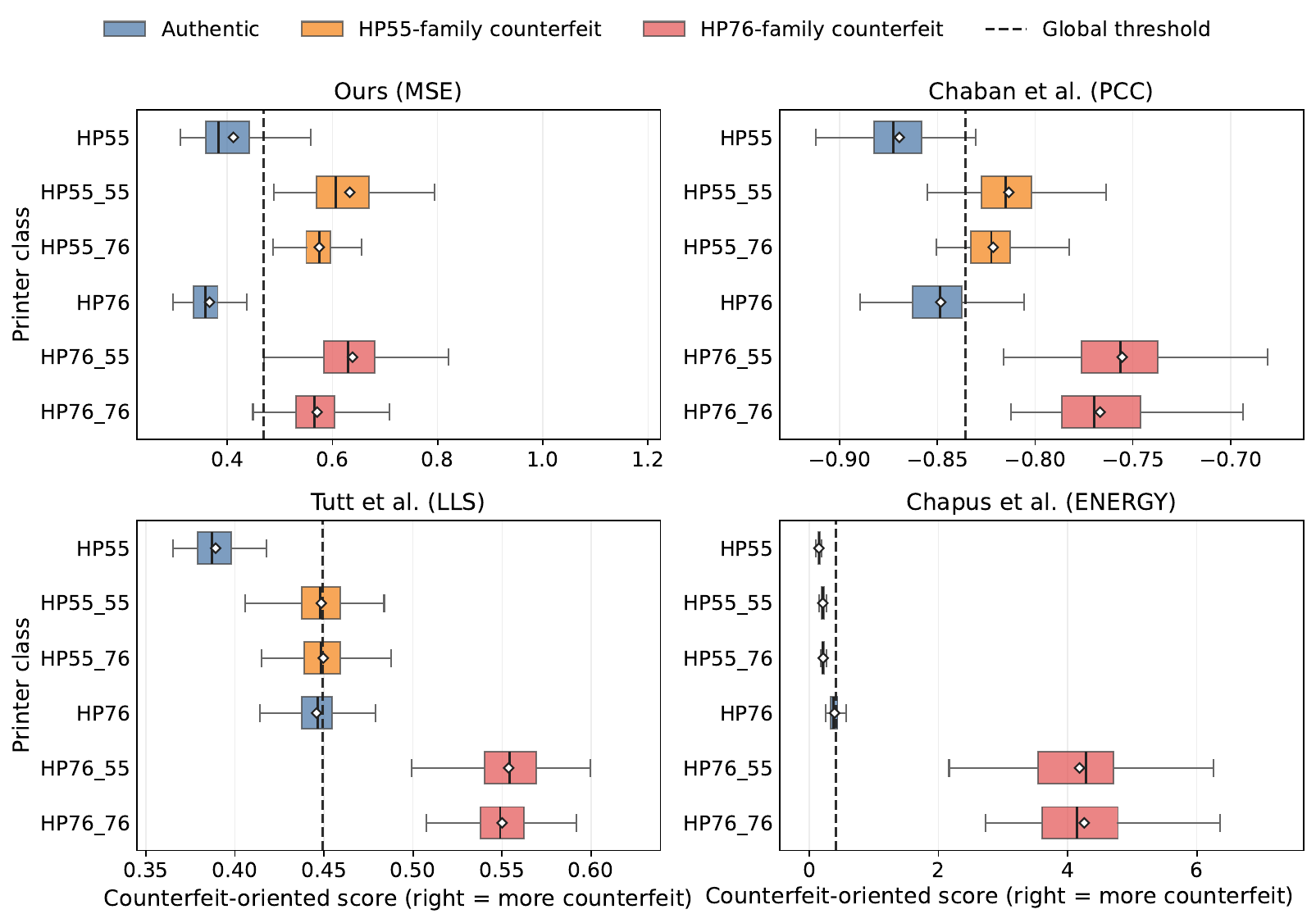}
    \caption{Score distributions under a global threshold ($\lambda=1\sigma$). Authentic samples should lie on the authentic side, while counterfeits on the anomalous side. The figure visualizes cross-class calibration, where authentic HP55 and HP76 samples should be comparable under the same decision rule.}
    \label{fig:score-distributions}
\end{figure}

However, strong per-family discrimination does not necessarily imply
strong multi-class authentication. In the multi-class setting, the same
score and threshold must be valid across all authentic P\&D classes.
\Cref{fig:score-distributions} illustrates why some baselines degrade in
this setting. For Chapus et al.~\cite{DBLP:conf/avss/ChapusCEB25}, authentic classes occupy different score ranges, so a threshold suitable for one authentic class does not transfer cleanly to the other. Similarly, Tutt et al.~\cite{DBLP:journals/tifs/TuttTCPBHV24} separates some authentic and counterfeit groups well locally, but its scores are not sufficiently calibrated across authentic classes for a single global threshold. Their lower combined AUROCs therefore reflect cross-class score miscalibration, rather than a complete inability to detect counterfeits in the per-family setting.

In contrast, our method produces compact authentic score distributions for both authentic classes under the same global threshold, while all counterfeit types are shifted toward higher anomalous scores.
This is the desired behaviour for multi-class normality modelling, where the model learns class-specific authentic manifolds, but the resulting
OOD scores remain comparable across classes. Chaban et
al.~\cite{DBLP:conf/wifs/ChabanPV24} is the strongest adapted baseline under
this criterion, confirming that learned reconstruction is a strong
foundation for CDP authentication. Nevertheless, its higher
fixed-threshold error indicates weaker global calibration at the selected
operating point.

\inlineheading{Operating-point behaviour}At the fixed authentic-only threshold, our method prioritizes low
false-acceptance rates. With MSE scoring, the mean false-acceptance rate
is $P_{\mathrm{fa}}=0.010$, while the mean missed-authentic rate is
$P_{\mathrm{miss}}=0.101$. The higher missed-authentic rate for HP55
relative to HP76 reflects greater within-class score variability for
HP55, causing some borderline authentic samples to exceed the global
threshold. In many authentication settings, this trade-off may be
preferable because false rejections can be routed to secondary
verification, whereas false acceptances directly compromise security.
The AUROC results complement this fixed-threshold analysis by showing
that the proposed score remains highly discriminative over the full range
of possible thresholds.

\begin{table}[t]
\centering
\caption{Combined AUROC and $P_{\text{err}}$ with $95\%$ bootstrap confidence intervals (5{,}000 template-level resamples).}
\label{tab:combined-bootstrap-ci}
\normalsize
\setlength{\tabcolsep}{18pt}
\begin{tabular}{lcc}
\hline
\textbf{Method} & \textbf{Combined AUROC [95\% CI]} & $\mathbf{P}_{\textbf{err}}$ \textbf{[95\% CI]} \\ \hline
Chaban et al.~\cite{DBLP:conf/wifs/ChabanPV24}  & 0.962 [0.952, 0.972] & 0.106 [0.085, 0.127] \\
Tutt et al.~\cite{DBLP:journals/tifs/TuttTCPBHV24}  & 0.886 [0.874, 0.898] & 0.239 [0.217, 0.261] \\
NCC  & 0.926 [0.909, 0.943] & 0.286 [0.256, 0.315] \\
Chapus et al.~\cite{DBLP:conf/avss/ChapusCEB25}  & 0.746 [0.744, 0.748] & 0.335 [0.316, 0.354] \\
Ours (MSE) & \textbf{0.975} [0.959, 0.988] & \textbf{0.055} [0.039, 0.073] \\ \hline
\end{tabular}
\end{table}

\inlineheading{Prompt Inference} As shown in \Cref{tab:inferred-prompts}, the diffusion classifier selects
the correct family prompt for $93.1\%$ of HP55 and $88.9\%$ of HP76
authentic samples (balanced accuracy $91.0\%$), and the same
family-consistent pattern carries over to counterfeits. Crucially, the
classifier does not degenerate to a single default prompt, and the authentication performance
is therefore realistic under test-time inferred conditioning without
access to ground-truth family labels.

\begin{table}[t]
\centering
\caption{Inferred test-time prompts. Entries are count (percentage).}
\label{tab:inferred-prompts}
\normalsize
\setlength{\tabcolsep}{25pt}
\begin{tabular}{l|l|c|c}
\hline
\textbf{Type} & \textbf{Class} & \textbf{HP55 prompt} & \textbf{HP76 prompt} \\ \hline
Authentic   & HP55    & 134 (93.1\%) & 10 (6.9\%)  \\
            & HP76    & 16 (11.1\%)  & 128 (88.9\%) \\ \hline
Counterfeit & HP55\_55 & 134 (93.1\%) & 10 (6.9\%)  \\
            & HP55\_76 & 136 (94.4\%) & 8 (5.6\%)   \\
            & HP76\_55 & 24 (16.7\%)  & 120 (83.3\%) \\
            & HP76\_76 & 26 (18.1\%)  & 118 (81.9\%) \\ \hline
\end{tabular}
\end{table}

\inlineheading{Effect of dual template masking}We isolate the contribution of the proposed
dual template masking strategy in \Cref{tab:masking-study}. Full-template reconstruction already achieves strong AUROC, but the score can still benefit from reducing direct reliance on visible template structure. A single checkerboard mask lowers $P_{\mathrm{miss}}$ but substantially increases $P_{\mathrm{fa}}$, indicating an unstable operating point. Random patch masking (covering one-fifth of the template) partially improves this trade-off but does not improve both
ranking and thresholded performance simultaneously. The complementary
dual-mask strategy achieves the best overall result, with AUROC
$=0.975$ and $P_{\mathrm{err}}=0.055$. By reconstructing and scoring
complementary withheld regions in two passes, the method reduces
dependence on visible binary structure while stabilizing the final score.
This improves sensitivity to subtle class-specific P\&D deviations under
authentic-class conditioning.

Overall, the results show that the proposed method addresses both
requirements of the task. It remains discriminative when authentic and
counterfeit samples are considered within each source printer family, and
it also provides a calibrated anomaly score suitable for a single
multi-class detector operating across multiple authentic P\&D classes.

\begin{table}[t]
\centering
\caption{Masking ablation using MSE.}
\label{tab:masking-study}
\normalsize
\setlength{\tabcolsep}{18pt}
\begin{tabular}{l|c|c|c|c}
\hline
\textbf{Strategy} & \textbf{AUROC} & $\mathbf{P}_{\text{err}}$ & $\mathbf{P}_{\text{miss}}$ & $\mathbf{P}_{\text{fa}}$ \\ \hline
No masking (full template)               & 0.968 & 0.070 & 0.135 & \textbf{0.005} \\
Single checkerboard         & 0.971 & 0.081 & \textbf{0.063} & 0.099 \\
Random patch            & 0.963 & 0.067 & 0.090 & 0.043 \\
\emph{Dual-mask} & \textbf{0.975} & \textbf{0.055} & 0.101 & 0.010 \\ \hline
\end{tabular}
\end{table}

\section{Conclusion}
\label{sec:conclusion}

This paper introduced a diffusion-based multi-class normality framework for unsupervised OOD detection. A single class-conditional diffusion model is trained only on authentic in-distribution classes and detects OOD samples through reconstruction error under authentic-class conditioning. We further introduced dual template masking, which hides complementary template regions and scores only withheld pixels to reduce reliance on visible binary structure.

The proposed method is applied to CDP authentication and outperforms traditional and adapted generative baselines on the Indigo $1\times1$ Base dataset. The results show that it supports both per-family authentic-versus-counterfeit discrimination and multi-class authentication with a single global decision rule across authentic P\&D classes. Future work will evaluate larger and more diverse printer and scanner configurations, additional conditioning signals, adaptive thresholding, and applications beyond CDP authentication.

\section{Acknowledgements}
\label{sec:acknowledgements}
This work was financed by the French National Research Agency (ANR), project TRUSTIT referenced under ANR-23-CE39-0002-01

\bibliographystyle{IEEEtran}
\bibliography{refs}

\end{document}